\documentclass{article}

\PassOptionsToPackage{numbers, compress}{natbib}
\usepackage[final]{nips_2018}

\usepackage[utf8]{inputenc} % allow utf-8 input
\usepackage[T1]{fontenc}    % use 8-bit T1 fonts
\usepackage{hyperref}       % hyperlinks
\usepackage{url}            % simple URL typesetting
\usepackage{booktabs}       % professional-quality tables
\usepackage{amsfonts}       % blackboard math symbols
\usepackage{nicefrac}       % compact symbols for 1/2, etc.
\usepackage{microtype}      % microtypography

\usepackage{hyperref}       % hyperlinks
\usepackage{url}            % simple URL typesetting
\usepackage{booktabs}       % professional-quality tables
\usepackage{amsfonts}       % blackboard math symbols
\usepackage{amsmath}
\usepackage{nicefrac}       % compact symbols for 1/2, etc.
\usepackage{microtype}      % microtypography
\usepackage{bm}
\usepackage{tabularx}
\usepackage{graphicx}
\usepackage{wrapfig}
\usepackage{caption}
\usepackage{subcaption}
\usepackage{float}
\usepackage{afterpage}

\newcommand{\matr}[1]{\bm{#1}}

\newcommand{\mW}[0]{\matr{W}}

\title{Object-oriented Targets for Visual Navigation using Rich Semantic Representations}

\author{
  Jean-Benoit Delbrouck\\
  Numediart \\
  University of Mons, Belgium\\
  \texttt{jean-benoit.delbrouck@umons.ac.be} \\
  \And
   St\'ephane Dupont\\
   Numediart \\
  University of Mons, Belgium\\
  \texttt{stephane.dupont@umons.ac.be} \\
}

\begin{document}

\maketitle
\vspace{-4mm}
\begin{abstract}  

  When searching for an object humans navigate through a scene using semantic information and spatial relationships. We look for an object using our knowledge of its attributes and relationships with other objects to infer the probable location. In this paper, we propose to tackle the visual navigation problem using rich semantic representations of the observed scene and object-oriented targets to train an agent. We show that both  allows the agent to generalize to new targets and unseen scene in a short amount of training time.
\end{abstract}
\vspace{-4mm}
\section{Introduction}

In this paper, we focus on the problem of navigating a space to find and reach a given object using visual input as well as rich semantic information about the observations. Given that the agent has a semantic knowledge of the world, the attributes defining the objects he sees are grounded into the environment. We would like the agent to use its capacity to understand the attributes and relationships between objects to find the quickest pathway towards a given target. We focus on two training aspects to achieve this task

Firstly, the agent is pre-trained to describe all aspects (object-wise) of a scene using natural language. This ability relies on a strong semantic understanding of a visual scene and guarantees a strong grasp of the environment. He is also able to output the localization of the described objects within the visual frame as well as the confidence about the given inference. We hope this strong ability gives the agent the semantic and spatial knowledge required for the task. Indeed, objects with similar utility or characteristics are usually close to each other (i.e. when looking for a stove, you might infer that it stands on cooking plates, as well as a keyboard lies close to a computer).

In practice, the previous statement doesn't always hold true. Some spatial relationships between objects are specific to certain space configurations. For example, in a living-room, a TV is usually in front of (and therefore close by) of sofa, even though the two objects don't share characteristics. But this spatial relation between these two objects is not especially true for, let's say, a bedroom. To tackle this second problem, our model is built in two distinct layers: one is dedicated to every scene (global knowledge about the world) and the other is scene-type specific (namely bedroom, bathroom, living-room or kitchen). No model layer is dedicated to a specific scene (an instance of a bedroom, for example) as opposed to prior work \citep{zhu2017target}, where a part of the model is reserved for each specific instance. In the latter contribution, the model tended to poorly transfer knowledge to unseen scene or objects.

For our experiments, we use the AI2-THOR framework \citep{ai2thor} which provides near photo-realistic 3D scene of house's room. The framework features enables the setup to follow a reinforcement setting. The state of the agent in the environment changes as the agent take actions. The location of the agent is known at each step and can be randomized for training or testing. We dispose of 20 rooms each containing 5 targets (or frame) distributed along the four scene-types: kitchen, living-room, bedroom and bathroom.

\section{Related work}
Since the emergence of virtual environments, a few works are related to ours. \citet{zhu2017target} proposed a deep reinforcement learning framework for target-driven visual navigation. As input to the model, only the target frame is given without any other priors about the environment. To transfer knowledge to unseen scene or object, a small amount of fine-tuning was required. We use their siamese network as baseline for our experiments. To adapt to unseen data, \citet{gupta2017cognitive} proposed a mapper and planner with semantic spatial memory  to output navigation actions. The use of spatial memory for visual question answering in such interactive environments has been explored by \citep{Gordon_2018_CVPR}. Visually grounded navigation instruction have also been addressed by a plurality of works \citep{Anderson_2018_CVPR, Yuiclr2018,hermann2017grounded}. The closest work related to ours is probably from \citet{yang2018}, they also use semantic priors for the virtual navigation task. Nevertheless, we differ in how the semantic representation is constructed. Their approach uses a knowledge graph built on Visual Genome \citep{Krishna:2017:VGC:3088990.3089101} where the input of each node is based on the the current state (or current visual frame) and the word vector embedding of the target. Our idea is rather based on natural language as well as object localization (detailed in section \ref{semantic_sec}). The task goal is also defined differently: their end state is reached when the target object is in the field of view and within a threshold of distance whilst we require the agent to go to a specific location.

\section{Architecture}
The following sections are structured as follows. First, we describe the main network architecture (\S\ref{baseline}), which is a siamese network (SN) used as a test-bed for our different experiments. The SN model is trained with random frames and then object-oriented frames (as explained in \S\ref{objoriented}) to set up two baselines. We improve the SN architecture with an semantic component (\S\ref{semantic_sec}). We explain how the semantic data are extracted from the agent observations (\S\ref{semantic-know}), and how we plug the component into our the baseline SN model (\S\ref{sem-comp}).

\subsection{Network Architecture} \label{baseline}

\begin{figure}[!h]
	\centering
	\includegraphics[scale=0.55]{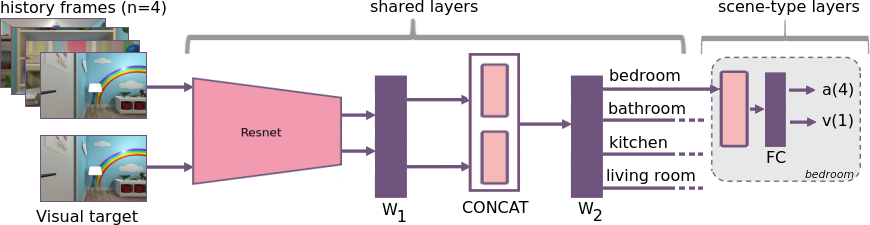}
	\caption{SN model. The right grey box can be seen as the policy layer, there is one for every scene-type. It means that each room instance will share its policy parameters with other instances of the same scene-type.}
\end{figure}

This section describes the siamese network (SN) model. First, four history frames and the target frames (which is tiled 4 times to match history frames input size) are passed through a ResNet-50 pre-trained on ImageNet \citep{imagenet_cvpr09}. Both inputs (history and targets) produces 2048-d features for each frame. We freeze the ResNet parameters during training. The resulting history and target 8196-d output vectors is linearly projected into the 512-d embedding space by the same matrix $\mW_1$. The CONCAT layer takes a 1024-d concatenated embedding and generates a 512-d joint representation with matrix $\mW_2$ that serves as input for each scene-type layer. This vector is subsequently passed through scene-type specific fully-connected layers  $f_{\text{s}}, s \in \{bathroom,bedroom,kitchen,livingroom\}$
, producing 4 policy outputs (actions) and a single value output. Matrices sizes of the models are $\mW_1 \in \mathbb{R}^{8196\times512}$, $\mW_2 \in \mathbb{R}^{1024\times512}$ and each $f_s$ contains two matrices $\mW_{s_1} \in \mathbb{R}^{512\times512}$ and $\mW_{s_2} \in \mathbb{R}^{512\times5}$. All matrices have their subsequent bias and use ReLu as activation function. \\

As the shared layers parameters are used across different room types, it can benefit from learning with multiple goals simultaneously. We then use A3C reinforcement learning model \citep{pmlr-v48-mniha16} that learns by running multiple copies of training threads in parallel and updates the shared set of model parameters in an asynchronous manner. The reward upon completion is set to 10. Each action taken decreases the reward by 0.01 hence favoring shorter paths to the target.
\subsection{Object-oriented targets} \label{objoriented}

We use the same data-set as used in \citet{zhu2017target} for comparison. It's composed of 20 rooms distributed along the four scene-types and for each room, five targets are chosen randomly. Our main assumption is that the use of semantic knowledge about objects improves the navigation task. As a first test-bed, we redefine the targets manually so that each target frame clearly contains an object. We make sure that an object picked appears on multiple other targets. To test this change, we make use of the model described in section \ref{baseline}. 

\begin{figure}[ht]
	\centering
	\includegraphics[scale=0.25]{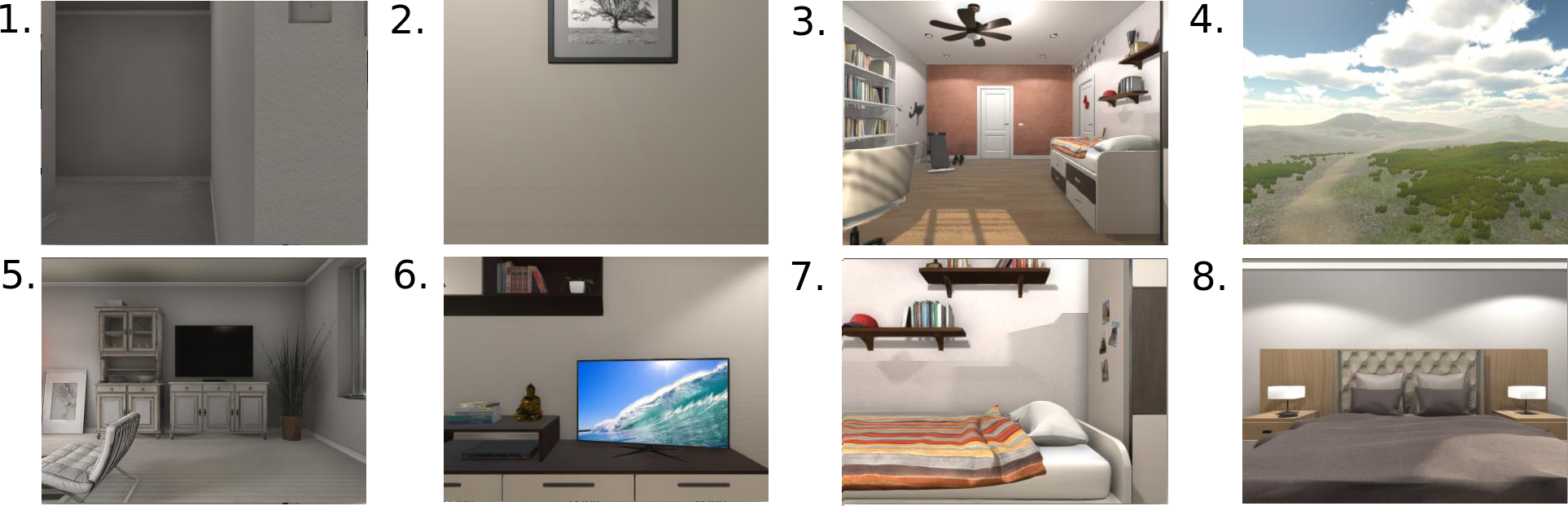}
	\caption{Handpicked differences between randomly sampled target (top) and object-oriented target (bottom) across 4 scenes.}
	\label{random_object_frames}
\end{figure}

Observations 1, 2, 3 and 4 are part of the random dataset. They are very hard for the agent to locate. Indeed, frame one and two are mostly walls and have no discriminating component. The frame number three is loaded with too many information while frame number four is the view through a window. Observations 5, 6, 7 and 8 are the corresponding frames (from the same room) in the object-oriented dataset. \\

Results in section \ref{results} will show that, after a certain amount of training, the SN model is able to learn with a moderate success to navigate to random targets. However, the training convergence (or the maximization of the reward) is much quicker when the model uses object-oriented targets. Moreover, providing object-oriented targets also improves the overall training so that it enables the agent to better generalize to other targets (random or object-oriented) and rooms. Both results are encouraging observations towards our intuition that the agent uses the objects and their relationship to navigate inside a room.

\subsection{Semantic architecture} \label{semantic_sec}
In this section, we explain how we collected data for our semantic knowledge database. We also show how we plug the semantics into the SN model presented in section \ref{baseline}.

\subsubsection{Semantic knowledge} \label{semantic-know}
To build the semantics data of the observations, we use DenseCap \citep{densecap}, a vision system to both localize and describe salient regions in images in natural language. DenseCap is trained on Visual Genome \citep{Krishna:2017:VGC:3088990.3089101}, an image data-set that contains dense annotations of objects, attributes, and relationships within each image. Every frame available in the 20 scenes are passed through the trained DenseCap model to generate our semantic knowledge database. An output for a frame is as follows:

\begin{figure}[ht]
	\centering
	\includegraphics[scale=0.50]{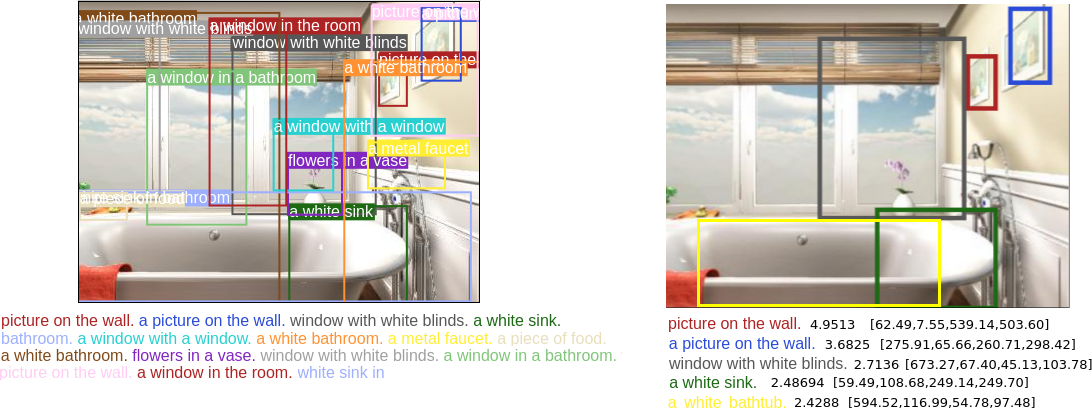}
	\caption{Output of DenseCap. We select the top-5 anchor boxes according to their confidence score.}
\end{figure}
For each anchor box, the model predicts a confidence score and four scalars ($[x\_min, y\_min, x\_max, y\_max]$) which represent the predicted box coordinates. DenseCap also includes a language model to generate a rich sentence for each box. Regions with high confidence scores are more likely to correspond
to ground-truth regions of interest, hence we keep 5 entries per frame having the top-5 confidence score. Even tough DenseCap already has a language model, we train an auto-encoder \footnote{\url{https://github.com/erickrf/autoencoder}} with all the selected sentences amongst all scenes as training set. It allows us to perform an in-domain training (4915 sentences with a 477 words vocabulary) and to define ourselves the feature-size for one sentence. Once the training has converged, we extract a 64-d vector per sentence. A semantic knowledge of a single frame is thus a concatenation of five vectors (one per anchor box) of 69 dimensions: 64-d for the sentence, 4-d for the anchor coordinates and one scalar for the confidence score.

\subsubsection{Semantic model}\label{sem-comp}

\begin{wrapfigure}{l}{0.5\textwidth}
  \begin{center}
    \includegraphics[width=0.50\textwidth]{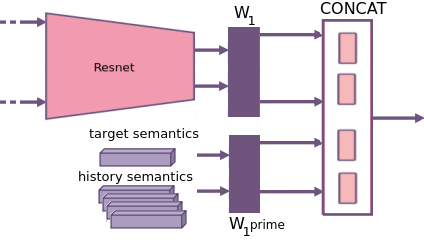}
  \end{center}
  \label{sem_model} 
  \vspace{-4mm}
\end{wrapfigure}
The semantic model (SSN) has two new inputs as shown in the figure on the left: the target semantics vectors and the history semantics. The target semantics is the semantic vector (of size 69 $\times$ 5 = 345) about the target frame and the history semantics is the concatenation of the semantic vectors of the past history frames (up to four). A new $\mW_{1^{\prime}}$ matrix encodes the semantic inputs and is of size $1380\times512$. The two visual inputs (visual history frames and visual targets) are similar to the SN model presented in section \ref{baseline}. \\
\\

\section{Results} \label{results}
We now detail our results in two parts: first we illustrate the convergence of the reward from the Siamese Network (SN) being trained on random targets and the object-oriented targets from section \ref{objoriented}. Finally, we describe the results of the semantic component compared with other models.

\begin{figure}[ht]
	\centering
	\includegraphics[scale=0.25]{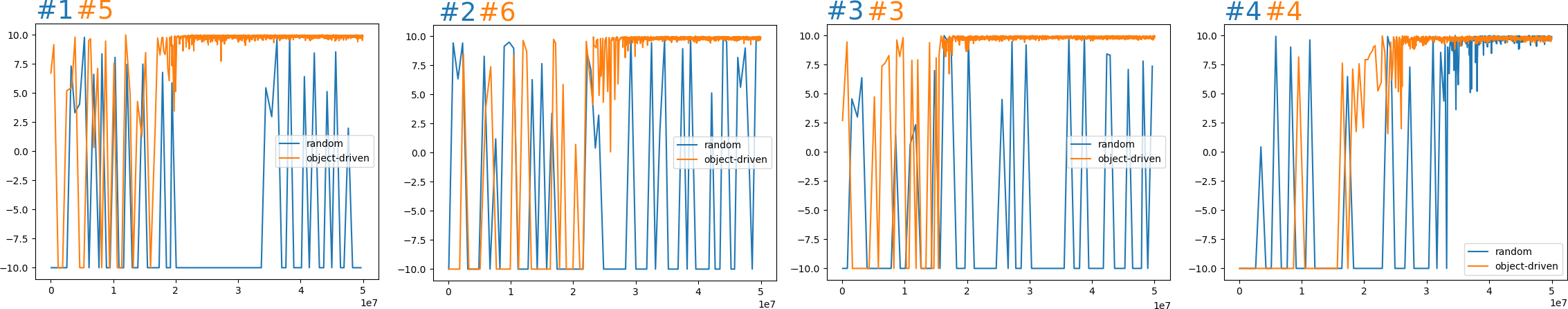}
	\caption{Evolution of the reward for the target observations ranging from 0 to 50 millions training frames.}
	  \label{convergence} 
\end{figure}

\textbf{SN baselines}  \quad Each graph of Figure \ref{convergence} contains two training curves (or reward evolution) of a target. The orange curve depicts an instance of the SN model trained on the object-oriented targets, and in blue, a instance of the SN model trained on random targets. The number above each graph refers to the target observation presented in Figure \ref{random_object_frames}. \\

As we see, the blue instance didn't converge for the random target \#1 and \#2. Indeed, both target are object-less: an empty corridor and a wall. The orange instance for target \#5 and \#6 has successfully converged. To compare their overall generalization ability, both blue and orange instances have been trained on the observation \#3 and \#4 (that are random). We notice that being trained on object-oriented target benefits the overall training: the orange instance converges quicker for random frames \#3 and \#4. \\

\textbf{Semantic}  \quad 
We evaluate the semantic model performances on two sub-tasks:
\begin{itemize}
    \item T1: generalization across targets (the scene is already seen, but not the targets)
    \item T2: generalization across scene (the scene has never been seen)
\end{itemize} 
We compare the semantic model (SSN) with the Siamese Network model (SN) model and previous work \citep{zhu2017target}. We consider an episode successful if the target is reached in less than 1000 actions. The experiments are conducted with 20 scenes (or rooms) instances with 5 object-oriented targets per instance, so 100 targets total. After training, each target is evaluated for 100 episodes and then averaged.

For task T1, we train the models for 5 millions frames shared across all scenes instances and then evaluate on unseen targets in one instance of each scene-type. We report accuracy for each scene type.

For task T2, the models are trained for 10 millions frames shared along all scenes instances except one instance per scene-type. The latter are used for evaluation.

Finally, we also evaluate our SSN model trained on top-semantic targets. Per instance, we redefine the five targets with the 5 frames that have the highest confidence scores from DenseCap (within each instance). A target to reach is now a observation the agent has the best semantic information about. We call this experiment SSN\_S.

\begin{table*}[!ht]
		\centering
		\hspace*{-1.15cm}
		\begin{tabular}{lcccccccc}
		
        \multicolumn{1}{c}{\bf T1}  &\multicolumn{2}{c}{\bf Bedroom}  &\multicolumn{2}{c}{\bf Bathroom}  &\multicolumn{2}{c}{\bf Kitchen} &\multicolumn{2}{c}{\bf Living}

			\\ \hline \\
			&E.L.& \%
			&E.L.&\%
			&E.L.&\%
			&E.L.&\%

			 \\
			R  
            & 963 & 6.6&  941 &9.4 &  988 & 6.4 & 968 & 4.8

            \\
			\citep{zhu2017target}  
            &903& 19.9 & 691 & 71.4 & 885& 23.9 & 921 & 16.2 
            \\
 			SN 
            &893& 22 & 611 & 78.2 & 890 & 22.1  & 901 & 18.1 
                        \\

            SSN
            &699& 61.9 & 358 & 92.4& 870 & 31.5 & 906 & 18.1 

            \\
			SSN\_S
            &619& 66.2 & 361 & 94.1 & 824 & 32.7 & 882 & 22.1 
            
            \\
            \\
                    \multicolumn{1}{c}{\bf T2}  &\multicolumn{2}{c}{\bf Bedroom}  &\multicolumn{2}{c}{\bf Bathroom} &\multicolumn{2}{c}{\bf Living}
 			\\ \hline \\
 		    &E.L.&\%
            &E.L.&\%
            &E.L.&\% & &
            \\
            R& 941 & 8.6&  928 &10.7 &  991 & 2.8 & &\\
            SN  &921& 9.6 & 861 & 12.4 & 966 & 4.1 & &\\
            SSN  &901& 10.4 & 811 & 17.4 & 982 & 4.1 & &\\
            SSN\_S &861& 14.5 & 798 & 19.1 & 954 & 4.7 & &

		\end{tabular}      
		\caption{Score table. R = Random, SN = Siamese Setwork model, SSN = semantic model, SSN\_S = semantic model trained on top-semantic targets, E.L. = mean episode length (rounded), \% = episode success rate.}
        \label{score-tabular}
	\end{table*}
	
\section{Conclusion} \label{concl}
For task 1, we see that the baseline Siamese Network (SN) using scene-type policies performs better overall than a scene-specific policy \citep{zhu2017target}. Also, our semantic model (SSN) is able to generalize to new targets within trained scene with a low amount of training frames. Importantly, training the semantic model on the top-semantic target (SSN\_N) instead of object-oriented target (SSN) offers an even better generalization to new targets. Task 2 is harder but nonetheless, the semantic model still comes ahead. It it possible that more training frames are required to transfer knowledge to new scenes.

\section{Acknowledgements}
This work was partly supported by the Chist-Era project IGLU with contribution from the Belgian Fonds de la Recherche Scientique (FNRS), contract no. R.50.11.15.F, and by the FSO project VCYCLE with contribution from the Belgian Waloon Region, contract no. 1510501. \\

\bibliographystyle{plainnat}
\bibliography{sample}

\end{document}